\documentclass[11pt]{article}

\usepackage[final]{acl}

\usepackage{times}
\usepackage{latexsym}
\usepackage{times}
\usepackage{latexsym}
\usepackage[T1]{fontenc}
\usepackage[utf8]{inputenc}
\usepackage{microtype}
\usepackage{inconsolata}
\usepackage{graphicx}
\usepackage{booktabs}
\usepackage{amsmath}
\usepackage{amssymb}
\usepackage{xcolor}
\usepackage{multirow}
\usepackage{array}
\usepackage{colortbl}
\usepackage{placeins}

\definecolor{best}{RGB}{198,239,206}
\definecolor{second}{RGB}{255,235,156}
\newcommand{\best}[1]{\cellcolor{best}\textbf{#1}}
\newcommand{\snd}[1]{\cellcolor{second}#1}

\newcommand{\method}{\textsc{Trident}}
\newcommand{\methodR}{\textsc{Trident-R}}
\newcommand{\methodS}{\textsc{Trident-S}}
\newcommand{\xkg}{\textsc{xKG}}

\usepackage[T1]{fontenc}

\usepackage[utf8]{inputenc}

\usepackage{microtype}

\usepackage{inconsolata}

\usepackage{graphicx}

\title{What the Reranker Sees: Multi-Aspect Page Annotation\\for Long-Document Multimodal Question Answering}

\author{
  \textbf{Guanchen Wu\textsuperscript{1}},
  \textbf{Jiayuan Ding\textsuperscript{2}},
  \textbf{Subhabrata Mukherjee\textsuperscript{2}}
  \textbf{Carl Yang\textsuperscript{1}}
\\
\\
  \textsuperscript{1}Department of Computer Science, Emory University, Atlanta, GA, USA
\\
  \textsuperscript{2}Hippocratic AI, Palo Alto, CA, USA
}

\begin{document}
\maketitle

\begin{abstract}
Long-document visual question answering (VQA) over documents of tens to hundreds of pages mixing text, tables, charts, and figures typically follows retrieve-then-read pipelines. In our setting, the bottleneck shifts from retrieval recall to reranker-side evidence selection: on MMLongBench-Doc, BGE-M3 reaches Recall@$20{=}0.86$ but only F1@$5{=}0.254$, and even the visual retriever ColPali reaches only F1@$5{=}0.332$---a text-only rerank LLM seeing only raw snippets misses table, chart, and layout evidence even when the upstream retriever encoded images. We propose \method{}, with two complementary components: \methodR{}, a retriever-agnostic LLM reranker that converts each candidate into an LLM-readable semantic record---visual caption, section path, entity tags, multi-axis concept hits, and a text snippet---then performs a single adaptive-$K$ rerank call; and \methodS{}, a generation-side module that prompts the VLM under topical, entity, and structural lenses before synthesis. On two long-document datasets, the annotation+rerank protocol substantially improves retrieval F1 across five heterogeneous pools, with every reranked pool exceeding the strongest adaptive-$K$ baseline PageIndex. An LLM rerank without the annotation barely changes first-hit ranking, indicating the lift comes from the structured annotation. \methodS{} targets open-ended synthesis questions by design, adding up to $6.6$ points in generation accuracy on these questions. The best \method{} configuration is the strongest downstream QA pipeline in our evaluation, with rankings consistent across two LLM judges ($\kappa{=}0.913$).
\end{abstract}

\section{Introduction}
\label{sec:intro}

\begin{figure}[t]
\centering
\includegraphics[width=0.95\columnwidth]{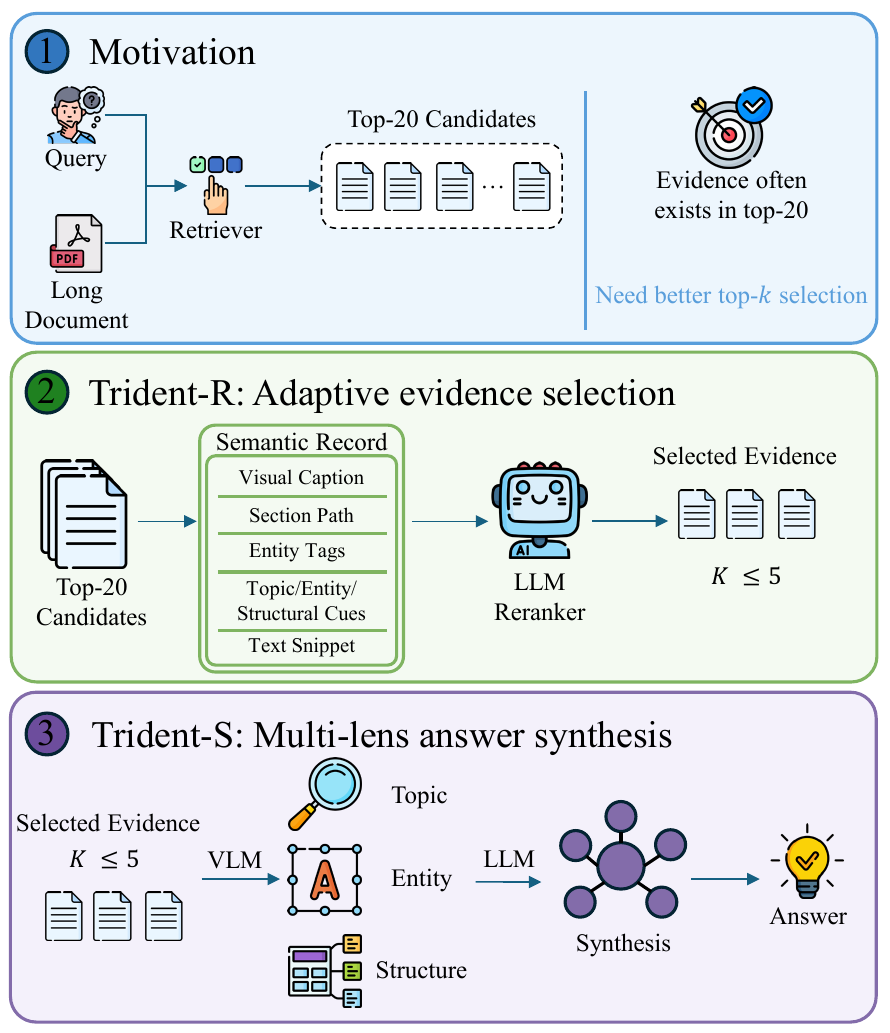}
\caption{
Overview of \method{}. 
Given a query and a long document, a retriever first returns a top-20 candidate pool where evidence pages often already appear. 
\methodR{} improves evidence selection by converting candidates into semantic records for adaptive LLM reranking, while \methodS{} synthesizes the final answer through topical, entity, and structural lenses.
}
\label{fig:intro}
\vspace{-7mm}
\end{figure}

Long-document visual question answering (VQA) requires answering natural-language questions over documents of tens to hundreds of pages that mix text, tables, charts, and figures. Although frontier Vision Language Models (VLMs) accept very long contexts in principle, feeding entire documents on every query is expensive and can suffer from position-induced degradation~\citep{liu2024lost}. Thus, retrieve-then-read pipelines~\citep{lewis2020retrieval} remain practical: a retriever selects a small set of pages, and a VLM answers the question conditioned on them. On MMLongBench-Doc, oracle evidence pages alone lift Large Vision Language Model (LVLM) F1 by $10$--$30$ points over full-document inputs~\citep{ma2024mmlongbench}, showing evidence selection is central to long-document VQA.

Where does evidence selection fail? In our setting, BGE-M3 ~\citep{chen2024bge} achieves $\mathrm{Recall}@20{=}0.86$ on MMLongBench-Doc but only $\mathrm{F1}@5{=}0.254$; the multimodal retriever ColPali~\citep{faysse2025colpali} improves this to $\mathrm{F1}@5{=}0.332$ but remains far from oracle. Evidence pages are typically present in the top-$20$ pool---the bottleneck is identifying the small evidence subset within $20$ surface-similar candidates.

A natural next step is to insert an LLM reranker between the top-$20$ pool and the $K{\le}5$ evidence selection, an approach used by RankGPT~\citep{sun2023chatgpt} and increasingly common in long-document RAG. However, when this reranker only sees raw text snippets of each candidate page, it does not improve first-hit ranking on either pool: on BGE-M3, MRR is essentially unchanged ($0.524 \to 0.523$); on ColPali, MRR actually \emph{drops} from $0.692$ to $0.598$, because the text-only LLM judgment overrides ColPali's stronger visual ranking signal. The reason is that long-document evidence is often locked in tables, charts, figures, and layout, all poorly represented in short text snippets. A rerank-stage LLM that only sees snippets is partially blind to exactly the evidence that matters most---ColPali's visual encoding occurs only at retrieval; the rerank LLM still consumes text-only candidate records and cannot access page images directly.

This rerank-stage blindness has been recognized: recent work addresses it by replacing the text LLM reranker with a multimodal LLM that directly sees page images~\citep{chen2025vlm, xu2025mm}. This is effective but expensive---every rerank query requires a full multimodal LLM call, and these rerankers typically need instruction-tuning or RL training on multimodal labels. We take an alternative architectural choice---a path used widely in industry RAG systems but underexplored in academic long-document QA: keep a cheap, training-free text-LLM reranker, and pre-compute a structured VLM-generated annotation once per document offline. Concretely, we convert each candidate into an LLM-readable semantic record---a VLM-generated visual caption surfaces table/chart/figure content, and structured fields (section path, entity tags, topic/entity/structure concept hits) expose pre-computed evidence signals. On the ColPali pool, attaching this annotation raises rerank F1 from $0.332$ to $0.581$. Consistent with the mechanism above, removing the caption hurts both pools but hurts BGE-M3 more than ColPali: a visual retriever partially compensates for the LLM's blindness at retrieval time, but does not eliminate the need for an explicit visual signal at rerank time.

Motivated by these observations, we propose \method{}, a long-document multimodal QA pipeline with two components addressing analogous bottlenecks at the rerank and generation stages. \methodR{} attaches the annotation above to each top-$20$ candidate and uses a single LLM call to select an adaptive set of $K{\le}5$ evidence pages. The same multi-aspect logic carries over to generation: a single VLM call must implicitly trade off attention across text, tables, and figures within each retrieved page, so we additionally propose \methodS{}, which prompts the VLM under topical, entity, and structural lenses before synthesizing a final answer. On MMLongBench-Doc, the annotation+rerank protocol applied to five heterogeneous candidate pools (BM25, BGE-M3, RRF, ColPali, and our multi-axis \xkg{}) raises retrieval F1 on all five, with every reranked pool exceeding the strongest adaptive-$K$ baseline PageIndex~\citep{zhang2025pageindex}; the improvement also holds on LongDocURL. \methodS{} targets open-ended synthesis questions by design, providing consistent gains on that workload while leaving extractive workloads to a standard single-call VLM. End-to-end, the best \method{} configuration is the strongest downstream pipeline in our evaluation, with rankings consistent across two LLM judges ($\kappa{=}0.913$).

\section{Related Work}
\label{sec:related}

\paragraph{Long-document VQA benchmarks and multimodal RAG systems.}
We evaluate on MMLongBench-Doc~\citep{ma2024mmlongbench} and LongDocURL~\citep{deng2025longdocurl}, two benchmarks targeting VQA over hundred-plus-page documents with multi-modal evidence (text, tables, charts, figures, layout); MMLongBench-Doc reports that GPT-4o-vision with full-document access lags VLMs given oracle evidence pages by ${\sim}30$ points. Recent benchmarks complement these along different axes: retrieval-focused MMDocIR~\citep{dong2025mmdocir} ($313$ long documents, notably finding VLM image descriptions outperform OCR text---consistent with our caption-dominance result), fine-grained evidence-selection MMDocRAG~\citep{dong2026benchmarking} (metrics beyond page-level F1), and difficulty-graded REAL-MM-RAG~\citep{wasserman2025real} (finance / technical). Multimodal RAG systems combine multi-modal retrievers with multimodal LLMs (M3DocRAG~\citep{cho2024m3docrag}), use hierarchical indexing with intra-page and cross-page chunks (MMRAG-DocQA~\citep{gong2025mmrag}), build chunk--query graphs (MLDocRAG~\citep{zhang2026mldocrag}), triage by document structure (PDFTriage~\citep{saad2024pdftriage}), or use VLM-generated structured JSON descriptions for tables and figures (MultiFinRAG~\citep{gondhalekar2025multifinrag}). \method{} differs by factoring the pipeline into a swappable Stage~1 retriever and a Stage~2 annotation $+$ rerank protocol we empirically isolate (\S\ref{sec:isolation}).

\paragraph{Document retrievers and adaptive-$K$ selection.}
Page-level retrievers span sparse (BM25), dense (BGE-M3~\citep{chen2024bge}; ColBERT-style late interaction~\citep{khattab2020colbert,santhanam2022colbertv2}; RRF fusion~\citep{cormack2009reciprocal}), and vision-aware variants (ColPali~\citep{faysse2025colpali} with PaliGemma plus patch-level MaxSim; DSE~\citep{ma2024unifying} with image-level dense embeddings); none condition on document-level structure. Adaptive-$K$ selection has been addressed by section-tree navigation (PageIndex~\citep{zhang2025pageindex}), hierarchical summarization (RAPTOR~\citep{sarthi2024raptor}), reflection-controlled retrieval (Self-RAG~\citep{asai2024self}), summary-based re-ranking (SimpleDoc~\citep{jain2025simpledoc}), and relevance clustering (AVIR~\citep{li2025avir}). Recent agentic approaches use iterative refinement (Doc-React~\citep{wu2025doc}), unified MLLM retrieval-generation (URaG~\citep{shi2026urag}), or multi-turn RL (MM-Doc-R1~\citep{lin2026mm}). These either rely on a single structural / semantic / visual signal or require end-to-end MLLM / agent training; \method{} is complementary---training-free, retriever-agnostic, and easily bolted onto any pipeline via multi-aspect annotation.

\paragraph{LLM rerankers with structured features.}
Retrieval-augmented generation~\citep{lewis2020retrieval} traditionally couples retriever and generator via end-to-end training; \method{} sits at the opposite end of the design space---modular and prompting-based, with multi-aspect annotation serving as a frozen abstraction layer pairable with any pool and any VLM without re-training. RankGPT~\citep{sun2023chatgpt} exposes raw candidate text to an LLM for re-scoring; recent work instead conditions LLM rerankers on \emph{structured per-candidate features}: CoRank~\citep{tian2025corank} (categories / sections / keywords for scientific retrieval), FinCARDS~\citep{zhou2026fincards} (schema fields for financial QA), KeyB2~\citep{li2024keyb2} (block selection), and AcuRank~\citep{yoon2026acurank} (uncertainty-aware adaptive listwise)---all text-only. \method{} differs in three ways: (i) we target \emph{long-document multimodal} QA, using VLM-generated visual captions as an LLM-readable interface for multimodal evidence; (ii) we test the interface via cross-pool isolation across five retriever paradigms, and show the gains transfer to the visual ColPali pool via asymmetric drop-caption and pool-invariant caption-only ablations (\S\ref{sec:isolation}, Appendix~\ref{app:detailed-ablations}); (iii) our generation-side analog \methodS{} extends the same principle to multi-view VLM prompting. We additionally expose multi-aspect metadata---concept hits across topic / entity / structure axes plus section path, visual caption, and entity tags---enabling annotation-level reasoning when surface text is repetitive among top-20 candidates. Knowledge graphs have been used for retrieval over textual corpora~\citep{gutierrez2024hipporag} and document QA~\citep{sanmartin2024kg}; \method{} differs by treating annotation as a page-level reranker-friendly abstraction layer, not as a graph-walk substrate.

\paragraph{Caption-based vs MLLM-based multimodal reranking.}
Two architectural families incorporate visual content into multimodal rerankers. \textbf{Caption-based approaches} convert page images to text via VLM descriptions and feed a text-only LLM reranker; this pattern is common in industry RAG systems (Haystack, NVIDIA RAG Blueprint, IBM Multimodal RAG) but academically underexplored under a ``loses visual details'' assumption. \textbf{MLLM-based approaches} let a multimodal LLM see page images directly, spanning zero-shot prompt-based rerankers~\citep{chen2025vlm,mortaheb2025re}, fine-tuned / RL-trained variants~\citep{xu2025mm,wasserman2025docrerank,lin2025mm}, and listwise / production rerankers~\citep{li2026qwen3}; all are effective but costly per query and typically require multimodal instruction-tuning or RL training. Concurrent work also retrieves over VLM-encoded pages directly~\citep{yu2025visrag}. \method{} systematically revisits the caption-based path for long-document multimodal QA and shows that with structured multi-aspect annotation, a text-only LLM reranker is competitive with or better than text-only / visual baselines at a fraction of per-query cost, while remaining drop-in compatible with any upstream retriever---tested via cross-pool isolation across five retriever paradigms (\S\ref{sec:isolation}).

\paragraph{VLM-augmented LLM QA and multi-view prompting.}
SCRA-VQA~\citep{zhang2025scra} converts single-image VQA into a caption $+$ LLM rerank pipeline, demonstrating that VLM captions can substantially augment LLM-based QA at the single-image scale; MPCAR~\citep{rahman2025mpcar} generates diverse complementary descriptions from multiple analytical perspectives and fuses them in a single-image LVLM reasoning prompt. \method{} extends this lineage to \emph{long-document} multimodal retrieval and generation: visual caption is identified as the strongest universal annotation field on the retrieval side; the three orthogonal lenses on the generation side (\methodS{}) are structured around our T / E / S decomposition axes and are followed by a format-aware LLM synthesis call. Crucially, we \emph{characterize} \methodS{}'s effective regime (substantial gains on open-ended synthesis questions only) rather than claim universal lift.

\section{Method}
\label{sec:method}

\begin{figure*}[t!]
\centering
\includegraphics[width=0.95\textwidth]{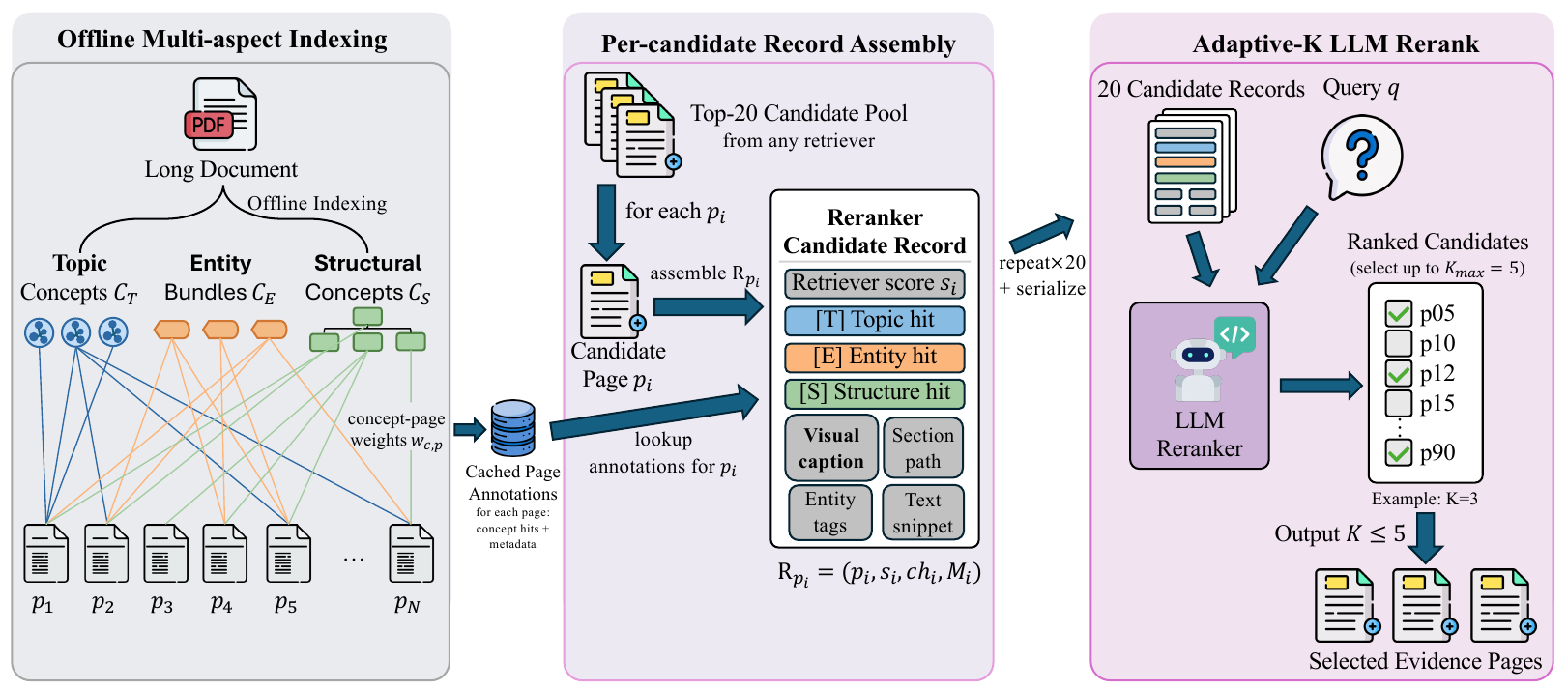}
\caption{Overall framework of TRIDENT-R.}
\label{fig:framework}
\end{figure*}

\subsection{Problem Formulation}
\label{sec:problem}

Let a document $D = \{p_1, \ldots, p_N\}$ contain $N$ pages of mixed text and visual content ($N$ may exceed $100$). Given a natural-language question $q$ with gold evidence set $E_q \subseteq D$ and gold answer $a_q$, the long-document VQA task is to produce an answer $\hat a$ approximating $a_q$. We decompose the task into retrieval and generation:
\begin{equation}
\hat P_q = \mathcal{R}(q, D),\ |\hat P_q| \le K_{\max};\quad
\hat a   = \mathcal{G}(q, \hat P_q),
\end{equation}
where $\mathcal{R}$ returns up to $K_{\max}$ evidence pages and $\mathcal{G}$ is a VLM. Following the reranker-interface bottleneck identified in \S\ref{sec:intro}, \method{} factors $\mathcal{R}$ into a candidate-retrieval stage and an annotation $+$ rerank stage:
\begin{equation}
\hat P_q = \mathcal{R}_2(q, \mathcal{R}_1(q, D),\ \Phi(D)),
\end{equation}
where $\mathcal{R}_1$ returns a top-$M{=}20$ pool, $\Phi(D)$ is a multi-aspect annotation (\S\ref{sec:method-index}), and $\mathcal{R}_2$ is an LLM reranker with $K_{\max}{=}5$ (\S\ref{sec:method-rerank}). $\mathcal{G}$ is then realized by either a single VLM call or a multi-view generator \methodS{} (\S\ref{sec:method-gen}). $\mathcal{R}_1$ is treated as a swappable candidate-pool source (\S\ref{sec:method-pool}), while $\mathcal{R}_2$ and the page annotation $\Phi$ are the focus of this work.

\subsection{\methodR{}-Index: Multi-Aspect Page Annotation}
\label{sec:method-index}

\methodR{}-Index is a function $\Phi(D) = (\mathcal{C}, \mathcal{M})$ that maps a document $D$ to a set of concepts $\mathcal{C}$ and per-page metadata records $\{\mathcal{M}_p\}_{p \in D}$. The concept set decomposes into three disjoint axes:
\begin{equation}
\mathcal{C} \;=\; \mathcal{C}_T \cup \mathcal{C}_E \cup \mathcal{C}_S
\end{equation}
where $\mathcal{C}_T$ contains topic concepts, $\mathcal{C}_E$ entity-bundle concepts, and $\mathcal{C}_S$ structural concepts. Each concept $c \in \mathcal{C}_X$ ($X \in \{T, E, S\}$) carries a name, an embedding $\mathbf{e}_c \in \mathbb{R}^{d}$ (where applicable), and a \emph{weighted membership}---a set of (page, weight) pairs $\{(p, w_{c,p}) : p \in D,\ w_{c,p} > 0\}$ where $w_{c,p} \in [0,1]$ reflects how strongly page $p$ contributes to concept $c$.

\paragraph{Topic concepts $\mathcal{C}_T$.}
We segment each page $p$ into overlapping text windows of $\leq 512$ tokens (stride $256$), denoted $W_p = \{w^{(1)}_p, \ldots, w^{(n_p)}_p\}$. All windows $\bigcup_p W_p$ are embedded by a sentence embedding model $\mathcal{E}_W$ and L2-normalized. We cluster the embedding set via mini-batch $K$-means with $K_T{=}\min(64, \tfrac{|W|}{8})$. Each resulting cluster $c \in \mathcal{C}_T$ records its centroid $\mathbf{e}_c$, member windows $W_c$, and per-page weight:
\begin{equation}
w_{c,p} \;=\; \frac{|W_c \cap W_p|}{|W_c|},
\quad \text{so that } \sum_p w_{c,p} = 1.
\end{equation}

\paragraph{Entity-bundle concepts $\mathcal{C}_E$.}
An extraction LLM $\mathcal{L}_{\text{ext}}$ extracts a flat set of typed entities $\mathcal{E}_p$ from each page---named entities, numbers, dates, monetary values, percentages, and table / figure references. We group co-occurring entities across pages using density-based clustering on a graph where entities are connected when they co-occur on at least one page, retaining bundles of size $\ge 2$. Each bundle $c \in \mathcal{C}_E$ stores its member entity set $\mathcal{E}_c$ and the pages on which $\mathcal{E}_c$ surfaces, with weight $w_{c,p}$ equal to the fraction of $\mathcal{E}_c$ entities appearing on $p$.

\paragraph{Structural concepts $\mathcal{C}_S$.}
We obtain a hierarchical section tree of $D$ from a public PDF structural parser $\mathcal{P}_{\text{sec}}$ (any document parser returning a section hierarchy can be substituted; we ground our choice in \S\ref{sec:exp-setup}). Each leaf section $s$ becomes a structural concept $c$ with member pages $\{p : p \in \text{span}(s)\}$ (uniform weights), and embedding $\mathbf{e}_c = \mathcal{E}_W(\text{summary}(s))$, where $\text{summary}(s)$ is the section summary returned by $\mathcal{P}_{\text{sec}}$.

\paragraph{Per-page metadata $\mathcal{M}_p$.}
In addition to the concepts above, we attach four per-page fields to each $p$: (a) a \emph{structural path} string (e.g.\ ``Item~8 / Consolidated Balance Sheets''), read off the section tree; (b) a \emph{visual caption} $\text{vc}_p$, generated by a captioning VLM $\mathcal{V}_{\text{cap}}$ prompted to describe the page's visible tables, figures, charts, and layout in $\leq 60$ words; (c) a flat \emph{entity-tag list} $\text{et}_p \subseteq \mathcal{E}_p$ retaining the top-15 most salient entities/numbers on the page; and (d) the page's raw text snippet (truncated to 600 characters).

\paragraph{Offline cost.} Index construction requires per-page LLM/VLM passes once per document and is cached on disk for reuse across all subsequent queries; full token-level accounting is in Appendix~\ref{app:cost}.

\subsection{\methodR{}: Adaptive-$K$ LLM Rerank}
\label{sec:method-rerank}

Given a query $q$, we first obtain a candidate page pool $P_q \subseteq D$ with $|P_q| \leq M = 20$ from one of the retrievers in \S\ref{sec:method-pool}, then \methodR{} performs a single LLM call that returns an adaptive-$K$ ordered subset $\hat P_q \subseteq P_q$ with $|\hat P_q| \leq K_{\max} = 5$.

\paragraph{Candidate payload.}
For each candidate page $p \in P_q$, we assemble a structured record $\mathcal{R}_p = (p,\ s_p,\ \text{ch}_p,\ \mathcal{M}_p)$ where $s_p$ is the underlying retriever score and $\text{ch}_p$ is a \emph{concept-hit list}
\begin{equation}
\begin{aligned}
\text{ch}_p \;=\; \big\{ &(X, c.\text{name}) : c \in \mathcal{C}_X, \\
&w_{c,p} > 0,\ X \in \{T, E, S\}\big\}
\end{aligned}
\end{equation}
collapsing all concept memberships involving $p$ into a typed list of $\leq 6$ tags (e.g.\ \texttt{T:Financial Statements}, \texttt{E:Nike}). Combined with $\mathcal{M}_p$, each candidate exposes \emph{five} discriminating signals to the reranker: concept hits, section path, visual caption, entity tags, and a truncated text excerpt.

\paragraph{Rerank step.}
We serialize $\{\mathcal{R}_p\}_{p \in P_q}$ as a JSON list and prompt a rerank LLM $\mathcal{R}_\theta$ (at $\tau{=}0$) with the question and candidate list, expecting a JSON response of the form:
\begin{equation}
\mathcal{R}_\theta(q, \{\mathcal{R}_p\}) \;=\; \big(\text{thinking},\ \hat P_q\big),
\quad |\hat P_q| \leq K_{\max}.
\end{equation}
The reranker may return fewer than $K_{\max}$ pages when the question's evidence is concentrated, yielding an adaptive $|\hat P_q|$. The verbatim rerank prompt is in Appendix~\ref{app:prompt}.

\paragraph{Why JSON-conditioned rerank?}
Compared to a fully-textual rerank prompt (e.g.\ RankGPT~\citep{sun2023chatgpt}), exposing structured per-candidate fields lets the LLM attend to specific evidence signals---visual caption, entity tags, concept memberships---rather than re-deriving them from the raw page text in its limited context. We empirically validate this design in \S\ref{sec:isolation} and isolate per-field contributions in \S\ref{sec:mechanism}.

\subsection{Candidate Pool Sources}
\label{sec:method-pool}

\methodR{} is agnostic to the source of $P_q$. We evaluate four pool choices, each returning the top-$M{=}20$ pages by score.

\paragraph{Single-signal text pools.}
\textbf{BM25} returns the top-20 pages by Okapi BM25 score over page text. \textbf{BGE-M3}~\citep{chen2024bge} returns the top-20 pages by cosine similarity between the query embedding and page embeddings (dim $1024$). \textbf{RRF}~\citep{cormack2009reciprocal} fuses BM25 and BGE-M3 rankings by Reciprocal Rank Fusion ($k{=}60$).

\paragraph{Vision pool.}
\textbf{ColPali}~\citep{faysse2025colpali} encodes each page image with PaliGemma and returns top-20 pages by late-interaction MaxSim.

\paragraph{Multi-axis pool (\xkg{}).}
We compose a multi-channel score from $\mathcal{C}_T, \mathcal{C}_E, \mathcal{C}_S$. A lightweight LLM router (Appendix~\ref{app:router}) first classifies $q$ into a class $\kappa \in \{\text{topical},\ \text{entity-level},\ \text{structural},\ \text{mixed}\}$ and extracts query entities $\mathcal{E}_q$. The concept-level scores per axis are:
\begin{align}
\text{score}_T(c, q) &= \mathrm{cos}(\mathbf{e}_c, \mathbf{e}_q),
  & c \in \mathcal{C}_T \\
\text{score}_E(c, q) &= \frac{|\mathcal{E}_c \cap_{\text{sub}} \mathcal{E}_q|}{|\mathcal{E}_c|},
  & c \in \mathcal{C}_E \\
\text{score}_S(c, q) &= \mathrm{cos}(\mathbf{e}_c, \mathbf{e}_q),
  & c \in \mathcal{C}_S
\end{align}
where $\cap_\text{sub}$ denotes loose substring matching to handle entity surface variants. Page-level scores are obtained by aggregating concept contributions weighted by membership and a class-conditional fusion $(\alpha_T^\kappa, \alpha_E^\kappa, \alpha_S^\kappa)$:
\begin{equation}
s^{\xkg{}}_p(q) \;=\; \sum_{X \in \{T,E,S\}} \alpha_X^\kappa
  \sum_{c \in \mathcal{C}_X} \text{score}_X(c, q) \cdot w_{c,p}.
\label{eq:fusion}
\end{equation}
The top-$M$ pages by $s^{\xkg{}}_p$ form the \xkg{} pool. Specific $(\alpha_T^\kappa, \alpha_E^\kappa, \alpha_S^\kappa)$ values for each $\kappa$, and a router-ablation comparison with uniform weights, are in Appendix~\ref{app:router}.

\subsection{\methodS{}: Multi-View Generation}
\label{sec:method-gen}

Given retrieved pages $\hat P_q$, \methodS{} prompts a VLM $\mathcal{V}$ three times in parallel under three lenses applied to the \emph{same} retrieved image set $\{\text{render}(p) : p \in \hat P_q\}$:
\begin{itemize}\setlength{\itemsep}{0pt}
  \item[\textbf{T:}] \emph{Topical lens.} ``What is each page about
        thematically? Which topics relate to the question?''
  \item[\textbf{E:}] \emph{Entity / value lens.} ``Extract the named entities,
        numerical values, dates, table cells, and figure labels visible on
        each page. Use verbatim values.''
  \item[\textbf{S:}] \emph{Structural lens.} ``Where does each page sit in the
        document organization (section title, captions, headers)?''
\end{itemize}
Producing three candidate answers $\hat a^T, \hat a^E, \hat a^S$. A fourth \emph{synthesis} call $\mathcal{S}$ merges them into a final answer:
\begin{equation}
\hat a \;=\; \mathcal{S}\big(q,\ \hat a^T,\ \hat a^E,\ \hat a^S\big),
\end{equation}
with format-aware preferences encoded in the synthesis prompt (preferring the E-view answer for numeric questions, the S-view for location questions, and consensus otherwise). When all three lenses agree the question is unanswerable, $\mathcal{S}$ returns ``Not answerable''---a calibration property we examine in \S\ref{sec:e2e}. Full prompts are in Appendix~\ref{app:mavs}.

\paragraph{Why three lenses?}
Long-document evidence is often heterogeneous within a single retrieved page (e.g., a financial-report page contains both narrative text \emph{and} a table). A single VLM call must implicitly trade off attention across modalities; by issuing three lens-conditioned prompts we explicitly induce the VLM to surface each evidence type, then defer the trade-off to a smaller textual synthesis call. This is conceptually analogous to multi-aspect annotation at generation time.

We treat \methodS{} as a question-format-conditioned extension; the empirical characterization of when it helps is in \S\ref{sec:e2e} and \S\ref{sec:longdocurl}.

\section{Experiments}
\label{sec:exp}

\subsection{Setup}
\label{sec:exp-setup}

\begin{table*}[t]
\centering\small
\setlength{\tabcolsep}{6pt}
\renewcommand{\arraystretch}{1.05}
\begin{tabular}{lcccccc}
\toprule
\textbf{Method} & \textbf{avg.\ $K$} & \textbf{Recall} & \textbf{Prec.} & \textbf{F1} & \textbf{MRR} & \textbf{nDCG} \\
\midrule
\multicolumn{7}{l}{\textit{Fixed-$K$ baselines (no rerank, $K{=}5$)}} \\
\midrule
BM25                & 5.00 & 0.525 & 0.162 & 0.231 & 0.448 & 0.437 \\
RRF                 & 5.00 & 0.556 & 0.172 & 0.246 & 0.524 & 0.496 \\
BGE-M3              & 5.00 & 0.572 & 0.178 & 0.254 & 0.524 & 0.502 \\
ColPali             & 5.00 & 0.739 & 0.236 & 0.332 & 0.692 & 0.675 \\
\midrule
\multicolumn{7}{l}{\textit{Adaptive-$K$, single-aspect signal}} \\
\midrule
PageIndex           & 2.99 & 0.644 & 0.442 & 0.480 & 0.569 & 0.583 \\
\midrule
\multicolumn{7}{l}{\textit{Adaptive-$K$, pool $+$ \method{} annotation $+$ rerank (ours)}} \\
\midrule
BM25 (sparse) $+$ ann.\ $+$ rerank   & 2.89 & 0.665 & 0.471 & 0.503 & 0.707 & 0.681 \\
RRF (hybrid) $+$ ann.\ $+$ rerank    & 3.78 & 0.748 & 0.454 & 0.511 & 0.751 & 0.728 \\
\xkg{} (graph) $+$ ann.\ $+$ rerank (\methodR) & \textbf{2.82} & 0.692 & 0.485 & 0.527 & 0.714 & 0.690 \\
BGE-M3 (dense) $+$ ann.\ $+$ rerank  & 2.89 & 0.726 & 0.499 & 0.546 & 0.744 & 0.723 \\
ColPali (visual) $+$ ann.\ $+$ rerank & 2.85 & 0.766 & 0.533 & 0.581 & 0.788 & 0.768 \\
\bottomrule
\end{tabular}
\caption{\textbf{Cross-pool isolation on MMLongBench-Doc} ($N{=}830$
aligned QAs; all reranked rows use GPT-4.1 with $K_{\max}{=}5$). The
same annotation $+$ rerank protocol improves sparse, hybrid, graph, dense,
and visual candidate pools.}
\label{tab:isolation}
\end{table*}

\begin{table}[t]
\centering\small
\setlength{\tabcolsep}{4pt}
\renewcommand{\arraystretch}{1.05}
\begin{tabular}{llcc}
\toprule
\textbf{Pool} & \textbf{Candidate view} & \textbf{F1} & \textbf{MRR} \\
\midrule
BGE-M3 & raw top-5 retrieval & 0.254 & 0.524 \\
BGE-M3 & text snippet $+$ LLM rerank & 0.374 & 0.523 \\
BGE-M3 & caption-only annotation $+$ rerank & 0.518 & 0.692 \\
BGE-M3 & full annotation $+$ rerank & 0.546 & 0.744 \\
\midrule
ColPali & raw top-5 retrieval & 0.332 & 0.692 \\
ColPali & text snippet $+$ LLM rerank & 0.423 & 0.598 \\
ColPali & caption-only annotation $+$ rerank & 0.554 & 0.744 \\
ColPali & full annotation $+$ rerank & 0.581 & 0.788 \\
\bottomrule
\end{tabular}
\caption{Mechanism ablations. LLM reranking with only raw text snippets
does not improve first-hit ranking on either pool (MRR flat on BGE-M3,
and decreasing on ColPali because the LLM judgment overrides ColPali's
visual ranking signal); the ranking lift requires the structured
annotation, and the visual caption alone recovers most of it on both
text and visual pools.}
\label{tab:mechanism}
\end{table}

\textbf{Datasets}: MMLongBench-Doc~\citep{ma2024mmlongbench} (1{,}082 QAs / 134 documents) and LongDocURL~\citep{deng2025longdocurl} (1{,}123 QAs / 200 documents). Retrieval metrics are reported on the $N{=}830$ answerable subset of MMLongBench-Doc. End-to-end accuracy is reported on a 300-QA stratified subset; the full sampling details are in Appendix~\ref{app:subset}.

\textbf{Baselines}: BM25, BGE-M3~\citep{chen2024bge}, RRF~\citep{cormack2009reciprocal}, ColPali~\citep{faysse2025colpali} at $K{=}5$ (fixed-$K$, no rerank), and PageIndex~\citep{zhang2025pageindex} adaptive-$K$ cookbook with GPT-4.1. Our cross-pool isolation pairs each retriever pool with the same \methodR{} annotation $+$ LLM rerank protocol.

\textbf{Module assignments}: We instantiate the abstract modules of \S\ref{sec:method} as follows---text embeddings ($\mathcal{E}_W$): \texttt{text-embedding-3-large}; extraction LLM $\mathcal{L}_{\text{ext}}$, rerank LLM $\mathcal{R}_\theta$, and synthesis LLM $\mathcal{S}$: GPT-4.1 at $\tau{=}0$; captioning VLM $\mathcal{V}_{\text{cap}}$: GPT-4o-mini; answering VLM $\mathcal{V}_{\text{ans}}$: GPT-4.1 (vision); structural parser $\mathcal{P}_{\text{sec}}$: the PageIndex API~\citep{zhang2025pageindex} used purely as a PDF section-tree extractor, independent of the PageIndex \emph{retrieval system} we benchmark against. The candidate JSON records produced by $\Phi$ are consumable by any instruction-tuned LLM; the annotation $+$ rerank protocol is therefore not tied to a specific model family. Claude and Gemini additionally serve as LLM-as-judge evaluators (Appendix~\ref{app:judges}).

\textbf{Metrics}: Retrieval is evaluated by page-level Recall, Precision, F1, MRR, and nDCG against gold evidence pages. End-to-end QA is evaluated by MMLongBench-Doc Generalized Accuracy and LLM-as-judge accuracy.

\textbf{PageIndex output}: PageIndex selects section nodes, which we expand to constituent pages and truncate at top-$5$ for apples-to-apples comparison. On MMLongBench-Doc this cap is rarely active ($18.6\%$ of queries); on LongDocURL it is more active ($49.7\%$).

\subsection{Cross-Pool Retrieval Isolation}
\label{sec:isolation}

Table~\ref{tab:isolation} tests whether the proposed interface works only for a particular retriever or transfers across candidate-pool distributions. The answer is consistent: applying the same annotation $+$ rerank protocol more than doubles raw retriever F1 for several pools (e.g., BGE-M3 $0.254 \to 0.546$) and raises all five pools above PageIndex. The visual ColPali pool is the most important case for our story: although ColPali already encodes page images, converting its candidates into LLM-readable annotations still raises F1 from $0.332$ to $0.581$. This indicates that the interface is not merely supplying missing visual information to text retrievers.

\subsection{Mechanism: What the Reranker Sees}
\label{sec:mechanism}

Table~\ref{tab:mechanism} isolates why the interface helps. The text-snippet-only rerank rows show that an LLM with raw page snippets improves F1 by shortening the output, but does not improve first-hit ranking on either pool: on BGE-M3 MRR is essentially unchanged ($0.524 \to 0.523$), and on ColPali MRR actually drops ($0.692 \to 0.598$) because the snippet-only LLM judgment overrides ColPali's stronger visual ranking signal. The ranking lift appears only once the reranker sees structured page annotation, and the visual caption alone is sufficient to recover most of it (BGE-M3 MRR $\to 0.692$, ColPali MRR $\to 0.744$ with caption-only; $\to 0.744 / 0.788$ with full annotation). Full per-field drop ablations (Appendix~\ref{app:detailed-ablations}) show the same pattern: removing visual caption hurts both pools, and the remaining topic/entity/structure axes add a small but nearly identical marginal gain ($+0.028$ / $+0.027$). Thus the caption is not just modality completion for text retrievers; it is a reranker-facing semantic representation that remains useful even when the upstream retriever is visual.

We keep detailed secondary analyses in the appendix to keep the main text focused on the reranker-interface claim. Appendix~\ref{app:detailed-ablations} reports full field ablations on BGE-M3 and ColPali; Appendix~\ref{app:router} reports \xkg{} router details; Appendix~\ref{app:k-by-qtype} reports the adaptive-$K$ distribution; and Appendix~\ref{app:stats} reports bootstrap, Wilcoxon, and McNemar tests.

\subsection{End-to-End QA as Supporting Evidence}
\label{sec:e2e}

\begin{table}[t]
\centering\scriptsize
\setlength{\tabcolsep}{2pt}
\renewcommand{\arraystretch}{1.05}
\resizebox{\linewidth}{!}{%
\begin{tabular}{lcccc}
\toprule
\textbf{Method} & \textbf{GPT J.} & \textbf{Claude J.} & \textbf{Gen Acc} & \textbf{Gen F1} \\
\midrule
BGE-M3 (Plain) & 44.7 & 44.7 & 43.5 & 41.3 \\
ColPali (Plain) & 48.3 & 46.3 & 46.4 & 45.9 \\
PageIndex $+$\methodS{} & 49.3 & 48.3 & 45.9 & 42.6 \\
BGE-M3 $+$ann.\ $+$rk $+$\methodS{} & 52.0 & 50.0 & 48.9 & 46.4 \\
ColPali $+$ann.\ $+$rk, Plain & 51.0 & 49.7 & 47.1 & 44.5 \\
\textbf{ColPali $+$ann.\ $+$rk $+$\methodS{}} & \best{53.7} & \best{52.7} & \best{51.2} & \best{48.9} \\
\methodR{} (\xkg{} pool, Plain) & 50.7 & 48.0 & 47.2 & 42.9 \\
\methodR{} (\xkg{} pool) $+$\methodS{} & 52.0 & 50.7 & 46.8 & 44.3 \\
\bottomrule
\end{tabular}%
}
\caption{End-to-end QA on the 300-QA MMLongBench-Doc subset (selected
pipelines; all values in \%). Full 14-pipeline results, including answer
accuracy and unanswerable abstention, are in Appendix~\ref{app:full-e2e}.}
\label{tab:e2e-main}
\end{table}

Table~\ref{tab:e2e-main} shows that the retrieval-side gain generally transfers to downstream QA. The ColPali pool with annotation and rerank is the best observed end-to-end configuration in this subset, and adding \methodS{} gives the highest Gen Acc and Gen F1 on MMLongBench-Doc. The full 14-pipeline table and inter-judge agreement are in Appendix~\ref{app:full-e2e} and Appendix~\ref{app:judges}; the pipeline ranking is consistent across the two judges ($\kappa{=}0.913$).

\subsection{LongDocURL Transfer Check}
\label{sec:longdocurl}

\begin{table}[t]
\centering\scriptsize
\setlength{\tabcolsep}{2pt}
\resizebox{\linewidth}{!}{%
\begin{tabular}{lcccccc}
\toprule
\textbf{Method} & \textbf{avg.\ $K$} & \textbf{Recall} & \textbf{Prec.} & \textbf{F1} & \textbf{MRR} & \textbf{nDCG} \\
\midrule
\multicolumn{7}{l}{\textit{Fixed-$K$ baselines (no rerank, $K{=}5$)}} \\
\midrule
BM25       & 5.00 & 0.623 & 0.191 & 0.281 & 0.616 & 0.568 \\
BGE-M3     & 5.00 & 0.649 & 0.205 & 0.298 & 0.622 & 0.575 \\
RRF        & 5.00 & 0.684 & 0.217 & 0.315 & 0.675 & 0.622 \\
ColPali & 5.00 & \textbf{0.750} & 0.235 & 0.343 & \textbf{0.763} & \textbf{0.700} \\
\midrule
\multicolumn{7}{l}{\textit{Adaptive-$K$}} \\
\midrule
PageIndex  & 4.03 & 0.598 & 0.302 & 0.366 & 0.430 & 0.463 \\
\methodR{} & 3.19 & 0.546 & \textbf{0.362} & \textbf{0.398} & 0.603 & 0.552 \\
\bottomrule
\end{tabular}%
}
\caption{LongDocURL retrieval comparison ($N{=}1{,}122$ QAs with non-empty
gold evidence). The retrieval-side trend transfers, although the full
cross-pool isolation study is conducted on MMLongBench-Doc.}
\label{tab:longdocurl}
\end{table}

Table~\ref{tab:longdocurl} provides an external transfer check on $N{=}1{,}122$ QAs. \methodR{} achieves the highest retrieval F1, showing that the retrieval-side trend transfers to LongDocURL. End-to-end results in Appendix~\ref{app:full-e2e} show the same pattern: ColPali plus annotation and rerank remains strongest.

\paragraph{When does \methodS{} help?} \methodS{} delivers consistent positive gains on MMLongBench-Doc's open-ended synthesis questions ($+1.7$ to $+6.6$ Gen Acc across all eight evaluated pipelines, with the largest absolute gain on weaker baselines---PageIndex Plain $\to{+}$\methodS{} adds $+6.6$). On LongDocURL's MCQ-style and short-extractive workloads, the three lenses converge on the same extracted answer, and \methodS{} is bypassed by design. We treat \methodS{} as a question-format-conditioned extension; an a-priori format router that gates \methodS{} invocation is a natural deployment refinement.

\section{Limitations and Conclusion}
\label{sec:limitations}

\paragraph{Limitations.} \methodR{} requires an offline per-document indexing pass ($\$0.256$/100 pages, amortized across queries; Appendix~\ref{app:cost}). The reranker uses GPT-4.1 here, but smaller or open-source LLMs are drop-in substitutes (Appendix~\ref{app:limits-extended}). The five-pool isolation is conducted on MMLongBench-Doc, with LongDocURL as an external transfer check (\S\ref{sec:longdocurl}). We use PageIndex only as a section-path parser; public PDF parsers such as PyMuPDF or GROBID could replace it. Finally, since most lift comes from visual captions, a caption-first variant is preferable when other fields are costly.

\paragraph{Conclusion.} We identify a reranker-interface bottleneck in long-document multimodal QA: evidence selection improves when candidate pages are converted into LLM-readable semantic annotations, even with multimodal retrievers. \methodR{} implements this as a retriever-agnostic annotation-based LLM reranker. Across five candidate-pool families on MMLongBench-Doc, the same interface substantially improves retrieval F1, and every reranked pool exceeds PageIndex. The gain on ColPali shows that the effect is not merely missing visual information for text retrievers. Mechanism ablations show that snippet-only LLM reranking barely changes first-hit ranking, while visual captions recover most of the full lift on both text and visual pools. \methodS{} is a secondary generation extension: useful for open-ended synthesis, but unnecessary for MCQ and short-extractive workloads. Overall, the results shift the focus from what the retriever scores to what the downstream reranker can see about each candidate page.

\bibliography{custom}
\clearpage

\appendix

\section{Additional End-to-End Results}
\label{app:full-e2e}

This appendix complements \S\ref{sec:e2e} and \S\ref{sec:longdocurl} with the full end-to-end QA tables on both benchmarks. Table~\ref{tab:e2e-full} reports all 14 evaluated pipelines on the 300-QA MMLongBench-Doc subset, including the answer-accuracy and unanswerable-abstention columns omitted from the main paper. Table~\ref{tab:longdocurl-e2e-full} reports the corresponding end-to-end results on the 300-QA LongDocURL stratified subset (all questions answerable), comparing \methodR{} against the strongest baseline ColPali $+$ annotation $+$ rerank, each with and without \methodS{}.

\begin{table*}[h]
\centering\small
\setlength{\tabcolsep}{5pt}
\renewcommand{\arraystretch}{1.05}
\begin{tabular}{lcccccc}
\toprule
\textbf{Method} & \textbf{GPT-4.1 J.} & \textbf{Claude J.} & \textbf{Gen Acc} & \textbf{Gen F1} & \textbf{Ans.\ Acc} & \textbf{Unans.\ Abst.} \\
\midrule
BM25 (Plain)             & 42.3 & 42.7 & 41.3 & 38.2 & 33.7 & 65.3 \\
BM25 $+$\methodS{}          & 45.0 & 44.3 & 44.8 & 43.8 & 39.7 & 61.1 \\
RRF  (Plain)             & 44.3 & 44.3 & 42.6 & 39.6 & 36.3 & 62.5 \\
RRF  $+$\methodS{}          & 44.7 & 43.3 & 43.9 & 41.7 & 39.3 & 58.3 \\
BGE-M3 (Plain)           & 44.7 & 44.7 & 43.5 & 41.3 & 37.9 & 61.1 \\
BGE-M3 $+$\methodS{}        & 46.0 & 45.7 & 43.9 & 43.4 & 41.1 & 52.8 \\
ColPali (Plain)          & 48.3 & 46.3 & 46.4 & 45.9 & 44.4 & 52.8 \\
ColPali $+$\methodS{}    & 50.0 & 49.0 & 49.2 & 48.1 & 47.2 & 55.6 \\
PageIndex (Plain)        & 42.7 & 43.0 & 41.3 & 37.2 & 29.7 & \best{77.8} \\
PageIndex $+$\methodS{}  & 49.3 & 48.3 & 45.9 & 42.6 & 40.6 & 62.5 \\
\midrule
BGE-M3 $+$ann.\ $+$rk, Plain         & 48.3 & 46.7 & 44.6 & 42.9 & 43.4 & 48.6 \\
BGE-M3 $+$ann.\ $+$rk, $+$\methodS{} & 52.0 & 50.0 & 48.9 & 46.4 & 48.1 & 51.4 \\
ColPali $+$ann.\ $+$rk, Plain        & 51.0 & 49.7 & 47.1 & 44.5 & 45.3 & 52.8 \\
\textbf{ColPali $+$ann.\ $+$rk, $+$\methodS{}} & \best{53.7} & \best{52.7} & \best{51.2} & \best{48.9} & \best{50.7} & 52.8 \\
\methodR{} (Plain)       & 50.7 & 48.0 & 47.2 & 42.9 & 41.9 & \snd{63.9} \\
\methodR{} $+$ \methodS{} & 52.0 & \snd{50.7} & 46.8 & 44.3 & 45.4 & 51.4 \\
\bottomrule
\end{tabular}
\caption{Full end-to-end QA results on the 300-QA MMLongBench-Doc subset
(all values in \%). \emph{J.}=LLM-as-judge accuracy; \emph{$+$ann.}=multi-aspect
annotation; \emph{$+$rk}=adaptive-$K$ rerank; \emph{Plain}=single multimodal
VLM call; \emph{$+$\methodS{}}=three-lens generation with synthesis.}
\label{tab:e2e-full}
\end{table*}

\begin{table}[h]
\centering\scriptsize
\setlength{\tabcolsep}{2pt}
\resizebox{\linewidth}{!}{%
\begin{tabular}{lcccc}
\toprule
\textbf{Method} & \textbf{GPT-4.1 J.} & \textbf{Claude J.} & \textbf{Gen Acc} & \textbf{Gen F1} \\
\midrule
\methodR{} (Plain)        & 51.3 & 47.7 & 42.6 & 47.3 \\
\methodR{} $+$\methodS{}  & 50.0 & 50.0 & 41.1 & 45.5 \\
\midrule
\textbf{ColPali$+$ann.\ $+$rk, Plain} & \best{68.7} & 65.7 & 56.0 & 58.0 \\
ColPali$+$ann.\ $+$rk, $+$\methodS{}   & 66.3 & \best{66.0} & \best{56.7} & \best{58.5} \\
\bottomrule
\end{tabular}%
}
\caption{End-to-end QA on the 300-QA LongDocURL stratified subset (all
values in \%; all questions are answerable). LongDocURL is dominated by
MCQ-style and short-extractive answers, so \methodS{} gives little average
benefit.}
\label{tab:longdocurl-e2e-full}
\end{table}

\section{\methodR{} Rerank Prompt}
\label{app:prompt}

The full rerank prompt sent to GPT-4.1 is shown below. Variables in angle brackets are filled at runtime. The candidate JSON list (\texttt{<candidates\_json>}) is the array of records described in \S\ref{sec:method-rerank}; \texttt{<query>} is the natural-language question; \texttt{<k>} is the maximum number of pages ($K{=}5$ throughout this work).

\begin{quote}\small\ttfamily
You are picking the top-K pages most likely to answer the question, from a
candidate list. Each candidate page comes with the document signals on it
AND a list of ``concept hits''---high-level concepts that pulled it into
the candidate pool. Pages hit by multiple distinct concepts are stronger
evidence.\\[0.4em]
Question: <query>\\[0.4em]
Candidate pages (<n> total):\\
<candidates\_json>\\[0.4em]
Reply as JSON: \{``thinking'': ``\ldots'', ``page\_list'': [<page\_id>, \ldots]\}\\
Pick the strongest evidence pages first. Return at most <k> pages. Return
only JSON.
\end{quote}

Each candidate record in \texttt{<candidates\_json>} has the form:
\begin{quote}\small\ttfamily
\{page\_id, retriever\_score, concept\_hits: [``T:<name>'', ``E:<name>'',
``S:<name>''], section, visual, entities, text\}
\end{quote}
The \texttt{retriever\_score} field holds the underlying Stage-1 pool
score (Okapi BM25 for BM25, cosine for BGE-M3, RRF fused rank for RRF,
late-interaction MaxSim for ColPali, or our multi-channel \xkg{} fusion
of Eq.~\ref{eq:fusion}); the field name is uniform across all pools so
the rerank prompt template is pool-agnostic. (In our released code the
field is currently named \texttt{graph\_score} for historical reasons;
the value semantics are as described.)

\section{\methodS{} Lens Prompts}
\label{app:mavs}

\methodS{} prompts the VLM three times in parallel with the three lenses below, then synthesizes a final answer with the synthesis prompt. All four calls share the same retrieved page images.

\paragraph{(T) Topical lens.}
\begin{quote}\small\ttfamily
Question: <query>\\[0.4em]
You are given <n> page images from a long document. Read them through the
\textbf{TOPICAL lens}: what is each page about thematically? Which topics
on which pages relate to the question? -- Identify the topical content of
each page. -- Focus on how the topics connect to the question.\\[0.4em]
Provide a clear, concise answer based on the topical analysis. If none of
the provided pages contain information sufficient to answer the question
through this topical lens, reply exactly: ``Not answerable''.
\end{quote}

\paragraph{(E) Entity / value lens.}
\begin{quote}\small\ttfamily
Question: <query>\\[0.4em]
\ldots Read them through the \textbf{ENTITY / VALUE lens}: focus on the
specific named entities, numerical values, dates, currencies, percentages,
table cells, and figure labels that appear on each page. -- Extract every
named entity, number, date, or specific value. -- Be precise---use
verbatim values from the images.\\[0.4em]
Provide a clear, concise answer focused on the specific entities and values
surfaced. If none of the provided pages contain the entities or values
needed to answer the question, reply exactly: ``Not answerable''.
\end{quote}

\paragraph{(S) Structural / section lens.}
\begin{quote}\small\ttfamily
Question: <query>\\[0.4em]
\ldots Read them through the \textbf{STRUCTURAL / SECTION lens}: focus on
where each page sits in the document's organization---section titles
visible on the page, headers, chapter / part markers, document structure,
captions of figures / tables. -- Identify each page's structural role.
-- Use captions of tables/figures and section paths.\\[0.4em]
Provide a clear, concise answer that leverages structural context. If the
structural context does not contain information needed to answer the
question, reply exactly: ``Not answerable''.
\end{quote}

\paragraph{Synthesis prompt.}
\begin{quote}\small\ttfamily
You are answering a long-document QA question. We elicited three candidate
answers from a vision-language model under three orthogonal analytical
lenses applied to the same K=<n> retrieved page images:\\[0.2em]
(T) Topical lens \quad (E) Entity / value lens \quad (S) Structural / section lens\\[0.4em]
Question: <query>\\[0.4em]
Candidate answers:\\{}
{}[T-view] <ans\_T>\\{}
{}[E-view] <ans\_E>\\{}
{}[S-view] <ans\_S>\\[0.4em]
Task: produce a single, clear, concise final answer.\\[0.4em]
Synthesis rules: -- If two or three lenses converge, prefer that answer.
-- If only one lens has a precise answer while others are vague, prefer
the specific one---unless contradicted. -- If question is numeric, prefer
E-view's value. -- If question is ``where / which section'', prefer S-view.
-- If question is ``what is the topic'', prefer T-view. -- Match the
question's expected answer format. Do not introduce information beyond
what the three candidate answers contain. Do not invent values. If all
three lenses agree the question is not answerable, reply exactly: ``Not
answerable''.
\end{quote}

\section{Query-Type Router}
\label{app:router}

The router prompts an LLM (GPT-4.1, $\tau{=}0$, JSON mode) to classify the question into one of four classes and to extract salient entities. The full prompt is:

\begin{quote}\small\ttfamily
You are routing a question against a multi-view document index that has
three concept types: topical (semantic content clusters), entity-bundle
(groups of co-occurring named entities/numbers), and section-derived
(structural sections).\\[0.3em]
Question: <query>\\[0.3em]
Output JSON with three fields:\\
\quad ``entities'': list of salient entities / key terms / numbers from the
question. Include named entities, technical terms, monetary amounts,
percentages, figure / table references. Drop generic stopwords.\\
\quad ``class'': one of ``entity-level'' (asks about a specific named
entity / number / figure), ``topical'' (asks for thematic content /
overview / explanation), ``structural'' (asks about location / section /
where in the document), or ``mixed'' if it spans multiple.\\
\quad ``rationale'': one short sentence explaining the class choice.\\[0.3em]
Reply with JSON only.
\end{quote}

The class label determines the channel-fusion weights $(\alpha_T, \alpha_E, \alpha_S)$ applied when aggregating topical / entity / structural concept scores into a candidate-pool score:

\begin{center}\small
\begin{tabular}{lccc}
\toprule
Class & $\alpha_T$ & $\alpha_E$ & $\alpha_S$ \\
\midrule
\texttt{topical}      & 0.60 & 0.20 & 0.20 \\
\texttt{entity-level} & 0.20 & 0.60 & 0.20 \\
\texttt{structural}   & 0.20 & 0.20 & 0.60 \\
\texttt{mixed}        & 0.40 & 0.30 & 0.30 \\
\bottomrule
\end{tabular}
\end{center}

Replacing this class-conditional router with uniform weights $(\tfrac{1}{3},\tfrac{1}{3},\tfrac{1}{3})$ causes a small but consistent $-0.003$ F1 drop on the cross-pool isolation setup (\S\ref{sec:isolation}, measured on the same $N{=}830$ aligned subset as Table~\ref{tab:isolation}), validating the router's contribution.

\section{300-QA Subset Construction}
\label{app:subset}

The 300-QA subset is drawn from MMLongBench-Doc with seed 42. Sampling emphasizes the long-document setting that motivates this work: the subset has median document length 72 pages, compared to 28 in the full set, aligning with our framework's focus on documents that exceed any current VLM's input window. Table~\ref{tab:subset-dist} reports the per-axis distribution match.

\begin{table}[t]
\centering\small
\setlength{\tabcolsep}{4pt}
\begin{tabular}{lccc}
\toprule
Axis (category) & Full \% & Subset \% & $\Delta$ pp \\
\midrule
\textit{Answer format} & & & \\
\quad Float                  & 14.8 & 20.7 & $+5.9$ \\
\quad Int                    & 26.4 & 24.0 & $-2.4$ \\
\quad List                   & 13.3 & 14.3 & $+1.0$ \\
\quad Str                    & 23.1 & 17.0 & $-6.1$ \\
\quad None (unans.)          & 22.4 & 24.0 & $+1.6$ \\
\midrule
\textit{Evidence source} & & & \\
\quad Chart                  & 16.1 & 14.2 & $-1.9$ \\
\quad Figure                 & 26.3 & 28.9 & $+2.6$ \\
\quad Layout                 & 10.9 &  8.0 & $-2.9$ \\
\quad Plain-text             & 26.9 & 28.4 & $+1.5$ \\
\quad Table                  & 19.8 & 20.4 & $+0.6$ \\
\midrule
\textit{Evidence pages} & & & \\
\quad 0 (unanswerable)       & 22.6 & 25.7 & $+3.1$ \\
\quad 1                      & 44.5 & 44.3 & $-0.1$ \\
\quad 2-3                    & 26.3 & 24.3 & $-2.0$ \\
\quad 4+                     &  6.7 &  5.7 & $-1.0$ \\
\midrule
\textit{Document length} & & & \\
\quad 0-30 pages             & 52.9 & 22.0 & $-30.9$ \\
\quad 31-60 pages            & 22.6 & 15.3 & $-7.2$ \\
\quad 61-120 pages           & 19.4 & 49.0 & $+29.6$ \\
\quad 121-200 pages          &  4.7 & 12.7 & $+8.0$ \\
\quad 200+ pages             &  0.5 &  1.0 & $+0.5$ \\
\bottomrule
\end{tabular}
\caption{Distribution match between the 300-QA subset and the full
MMLongBench-Doc dataset. Most axes (Evidence Source, Evidence Pages,
Unanswerable rate) match within $\pm 3$pp. Answer format is moderately
biased toward Float ($+5.9$pp) and away from Str ($-6.1$pp); document
length is intentionally biased toward longer documents (median 72 vs.\ 28
pages), aligning with our framework's long-document focus.}
\label{tab:subset-dist}
\end{table}

The LongDocURL subset of 200 QAs (\S\ref{sec:longdocurl}) is sampled analogously, stratified by document length, answer format, and evidence source. Released QA-id lists for both subsets accompany the code release.

\section{Bootstrap CIs, Wilcoxon and McNemar tests}
\label{app:stats}

\paragraph{Retrieval (N=830).}
We report (i) non-parametric 95\% bootstrap CIs on per-QA F1 ($B{=}10{,}000$ resamples drawn with replacement from the 830 aligned QAs); and (ii) pairwise Wilcoxon signed-rank tests on per-QA F1 against \methodR{} (xKG pool). We prefer Wilcoxon on F1 over McNemar on STRICT for retrieval because methods have different effective $K$ (5 for fixed-$K$ baselines; mean $K \in [2.7, 3.0]$ for adaptive-$K$ methods), which would structurally favor the larger-$K$ methods. CIs cluster tightly within $\pm 0.024$. All five raw baselines are significantly worse than \methodR{} ($p < 0.001$). Among the $+$ann.\ $+$ rerank variants, RRF and BGE-M3 swaps are statistically indistinguishable from \methodR{} ($p{=}0.077$ and $0.055$), the BM25 swap is borderline ($p{=}0.026$), and the ColPali swap is significantly higher ($p < 0.001$). Taken together, all five $+$ann.\ $+$ rerank rows in Table~\ref{tab:isolation} fall within the $[0.503, 0.581]$ F1 band without pairwise dominance (except ColPali), validating the pool-agnostic claim.

\paragraph{End-to-end (N=300).}
For end-to-end Gen Acc on the 300-QA MMLongBench-Doc subset we use McNemar on a binary correct-score ($\ge 1$) outcome per QA, with continuity correction. None of the pairwise comparisons between top-tier pipelines (ColPali$+$ann.\ $+$\methodS{}, \methodR{}$+$\methodS{}, BGE-M3$+$ann.\ $+$\methodS{}, PageIndex$+$\methodS{}) reach $p < 0.05$ at $N{=}300$, reflecting the small sample size; however the ranking is consistent across all three LLM judges (Appendix~\ref{app:judges}, inter-judge $\kappa{=}0.913$). Full pairwise results are released with our code.

\section{Case Study Records}
\label{sec:case-study}
\label{app:cases}

We list the full records for the three case-study QAs in \S\ref{sec:case-study}. All extracted answers are the GPT-4o-extracted short forms after the MMLongBench-Doc official evaluation pipeline. ``Score'' is the rule-based Generalized Accuracy score (1.0 = correct, 0.0 = wrong).

\paragraph{Case 1: Chart axes.}
Doc \texttt{2401.18059v1.pdf}; Q: \emph{What are the horizontal and vertical
axis of Figure~3 respectively?} Gold: \texttt{[Context Length, F1]}.
\begin{itemize}\setlength{\itemsep}{0pt}
  \item \methodR{}: retrieved pages $\{6, 5\}$, ext.\ ``\texttt{[Context
        Length, F1]}'', score 1.0.
  \item PageIndex: retrieved $\{3\}$, ext.\ ``\texttt{Not answerable}'',
        score 0.0.
  \item BGE-M3 raw: retrieved $\{5, 2, 16, 19, 22\}$, ext.\
        ``\texttt{Not answerable}'', score 0.0.
\end{itemize}

\paragraph{Case 2: Activision R\&D ratio.}
Doc \texttt{ACTIVISIONBLIZZARD\_2019\_10K.pdf}; Q: \emph{What is R\&D to asset
ratio for Activision Blizzard in FY 2019?} Gold: \texttt{5.03\%}.
\begin{itemize}\setlength{\itemsep}{0pt}
  \item \methodR{}: retrieved $\{70, 69\}$, ext.\ ``\texttt{0.0503}'',
        score 1.0.
  \item PageIndex: retrieved $\{37, 62, 67, 68, 69\}$, ext.\ ``\texttt{Not
        answerable}'', score 0.0.
  \item BGE-M3 raw: retrieved $\{47, 29, 92, 174, 175\}$, ext.\ ``\texttt{Not
        answerable}'', score 0.0.
\end{itemize}

\paragraph{Case 3: Calibrated refusal.}
Doc \texttt{2311.16502v3.pdf}; Q: \emph{In which image type does GPT-4o
demonstrate least proficiency \ldots?} Gold: \texttt{Not answerable}.
\begin{itemize}\setlength{\itemsep}{0pt}
  \item \methodR{}: retrieved $\{21, 111, 26\}$, ext.\ ``\texttt{Not
        answerable}'', score 1.0.
  \item PageIndex: retrieved $\{110, 111, 112\}$, ext.\ ``\texttt{Chemical}'',
        score 0.0.
  \item BGE-M3 raw: retrieved $\{112, 7, 26, 110, 2\}$, ext.\
        ``\texttt{Music}'', score 0.0.
\end{itemize}

\section{Detailed Cost Breakdown}
\label{app:cost}

Token-level cost accounting for \methodR{}-Index, measured empirically on the 134-document MMLongBench-Doc corpus (6{,}492 pages, mean $103$ text tokens per page, $40$ caption tokens per page) at public OpenAI 2025 rates (\$2.00/\$8.00 per 1M input/output tokens for gpt-4.1; \$0.15/\$0.60 for gpt-4o-mini; \$0.13 per 1M tokens for text-embedding-3-large; image tokens $\approx 1500$ per full page at $\text{ZOOM}{=}2.0$).

\begin{table}[h]
\centering\scriptsize
\setlength{\tabcolsep}{2pt}
\resizebox{\linewidth}{!}{%
\begin{tabular}{lrrr}
\toprule
Component (per 100-page doc) & Model & Calls & USD \\
\midrule
Visual caption           & gpt-4o-mini       & 100 & 0.028 \\
Entity extraction        & gpt-4.1           & 100 & 0.201 \\
Page embedding           & emb-3-large       & 100 & 0.001 \\
Concept naming           & gpt-4.1           &  20 & 0.026 \\
\midrule
\textbf{Offline total}   &                   &     & \textbf{0.256} \\
\midrule
\multicolumn{4}{l}{\textit{Online per-query (avg $K{=}2.85$):}} \\
Stage 2 LLM rerank       & gpt-4.1           &   1 & 0.010 \\
Plain VLM generation     & gpt-4.1 (vision)  &   1 & 0.012 \\
\methodS{} (3 lens $+$ synth) & gpt-4.1 (vision)  &   4 & 0.039 \\
\bottomrule
\end{tabular}%
}
\caption{Empirical token-level cost breakdown. Entity extraction
dominates offline indexing ($78\%$); concept naming and visual
captioning contribute $\sim 10\%$ each; page embedding is negligible
($1\%$).}
\label{tab:cost-breakdown}
\end{table}

The reported $\$0.256$ per 100-page document is within the $\sim \$0.30$ range stated in the main text and is conservative for dense documents: doubling per-page text tokens (to $\sim 200$) raises total cost to approximately $\$0.36$. Per-query cost is dominated by the multimodal VLM call: Plain generation is $\$0.012$, \methodS{} multi-view is $\$0.039$ (3 lens calls $+$ 1 synthesis).

\section{Detailed Field Ablation on Two Pools}
\label{app:detailed-ablations}

We complement \S\ref{sec:isolation} with the full per-field ablation on both the BGE-M3 (text-only) and ColPali (visual) pools. Each row removes exactly one annotation field from the rerank candidate JSON.

\begin{table}[h]
\centering\scriptsize
\setlength{\tabcolsep}{2pt}
\resizebox{\linewidth}{!}{%
\begin{tabular}{lcccc}
\toprule
Configuration & F1 & $\Delta$F1 & Recall & Prec.\ \\
\midrule
\multicolumn{5}{l}{\textit{BGE-M3 pool (text retriever, $\mathrm{F1}_{\text{full}}{=}0.546$)}} \\
$+$ full annotation        & 0.546 & ---     & 0.726 & 0.499 \\
\quad drop visual\_caption & 0.416 & $-0.130$ & 0.657 & 0.354 \\
\quad drop concept\_hits   & 0.515 & $-0.032$ & 0.688 & 0.477 \\
\quad drop entities        & 0.507 & $-0.039$ & 0.701 & 0.455 \\
\rowcolor{gray!12}
\quad \textbf{caption-only} (only visual\_caption) & \textbf{0.518} & $-0.028$ & 0.669 & 0.484 \\
\midrule
\multicolumn{5}{l}{\textit{ColPali pool (visual retriever, $\mathrm{F1}_{\text{full}}{=}0.581$)}} \\
$+$ full annotation        & 0.581 & ---     & 0.766 & 0.533 \\
\quad drop visual\_caption & 0.498 & $-0.083$ & 0.715 & 0.444 \\
\quad drop concept\_hits   & 0.593 & $+0.013$ & 0.751 & 0.556 \\
\quad drop entities        & 0.588 & $+0.007$ & 0.773 & 0.542 \\
\rowcolor{gray!12}
\quad \textbf{caption-only} (only visual\_caption) & \textbf{0.554} & $-0.027$ & 0.723 & 0.516 \\
\bottomrule
\end{tabular}%
}
\caption{Per-field annotation ablation on two retrievers ($N{=}830$ aligned, $K_{\max}{=}5$). Standard rows: remove one field at a time. \colorbox{gray!12}{Caption-only rows}: keep only the visual caption (drop concept\_hits, section path, entity tags, text snippet). Caption alone achieves ${\sim}90\%$ of the full annotation lift on both pools ($-0.028$ / $-0.027$ from full), with the gap remarkably pool-invariant. Combined with the asymmetric drop-caption result ($-0.130$ on BGE-M3 vs.\ $-0.083$ on ColPali; $36\%$ smaller on visual pool), this rules out pure modality completion as the sole mechanism: the lift stems from structuring page evidence as LLM-readable input, with visual caption acting as a \emph{semantic interface} for the text-only LLM reranker (effective regardless of whether the upstream retriever is visual or textual).}
\label{tab:field-ablation}
\end{table}

\section{Adaptive-$K$ Distribution by Question Type}
\label{app:k-by-qtype}

We bin the rerank-selected $K$ by MMLongBench-Doc question type (using
the \texttt{evidence\_sources} field) across four annotated-rerank pools.

\begin{table}[h]
\centering\small
\setlength{\tabcolsep}{4pt}
\begin{tabular}{lcccc}
\toprule
Question type & BM25 & BGE-M3 & ColPali & \methodR{} \\
              & $+$ann & $+$ann & $+$ann & (xKG)$+$ann \\
\midrule
Pure-text     & 2.95 & 3.08 & 3.11 & 2.94 \\
Figure        & 2.64 & 2.61 & 2.60 & 2.71 \\
Table         & 2.90 & 2.90 & 2.98 & 2.75 \\
Chart         & 2.81 & 2.77 & 2.55 & 2.59 \\
Layout        & 2.84 & 2.78 & 2.97 & 2.75 \\
Multi-source  & 3.01 & 2.98 & 2.95 & 2.97 \\
Unanswerable  & 2.69 & 2.72 & 2.67 & 2.62 \\
\bottomrule
\end{tabular}
\caption{Mean adaptive $K$ per question type, by candidate pool
($K_{\max}{=}5$). Two patterns: (1) $K$ is highly stable
\emph{across pools} (range $2.55$--$3.11$), suggesting that the
reranker's $K$ choice is driven by the question's evidence requirement,
not by which retriever fed the pool. (2) $K$ varies
\emph{by question type}: Pure-text questions request the most pages
($\bar K \approx 3.0$, $K{=}5$ cap rate $\sim 30\%$);
Figure / Chart questions request fewer ($\bar K \approx 2.6$,
$K{=}1$ rate $> 30\%$ on Figure), consistent with visual evidence
being concentrated on individual pages.}
\label{tab:k-by-qtype}
\end{table}

\section{Visual Caption QC: Substring and N-gram Overlap}
\label{app:caption-leakage}

To address the concern that VLM-generated captions might leak the gold
answer text and thereby trivialize Stage 2 reranking, we audit
$200$ randomly sampled answerable QA / candidate-page pairs ($N{=}238$
caption-page records after dropping ``Text-only page.'' captions).

\begin{itemize}\setlength{\itemsep}{0pt}
  \item \textbf{Exact substring} (gold answer string contained verbatim
        in caption): $22 / 238 = 9.2\%$.
  \item \textbf{Bigram Jaccard} $\geq 0.3$ (significant sentence-level
        overlap): $0 / 238 = 0.0\%$.
  \item \textbf{Unigram coverage} $\geq 0.7$ of gold tokens in caption:
        $32 / 238 = 13.4\%$.
\end{itemize}
The $0\%$ bigram match indicates no caption near-verbatim paraphrases
the answer sentence; the $9.2\%$ substring rate reflects captions
correctly encoding visual content (e.g.\ chart legend colours such as
``Blue'', or diagram labels such as ``Elastic Compute Service'') that
happen to coincide with the gold answer --- the intended behaviour of
caption-as-visual-evidence rather than answer leakage.

\section{Inter-Judge Agreement}
\label{app:judges}

We use two independent LLM judges (GPT-4.1 and Claude Sonnet 4.5) to
score the same 300-QA end-to-end gen outputs from each of the $14$
pipelines in Table~\ref{tab:e2e-full}. Both judges use the identical
$3$-level rubric (correct / partial / wrong) and the same MMLongBench-Doc
gold answers. Table~\ref{tab:inter-judge} reports per-pipeline pairwise
agreement and Cohen's $\kappa$.

\begin{table}[h]
\centering\small
\setlength{\tabcolsep}{4pt}
\begin{tabular}{lcc}
\toprule
Pipeline & Agree.\ & Cohen $\kappa$ \\
\midrule
BM25 (Plain / MAVS)              & $94.0$ / $96.7$ & $0.89$ / $0.94$ \\
BGE-M3 (Plain / MAVS)            & $94.3$ / $97.3$ & $0.90$ / $0.95$ \\
ColPali (Plain / MAVS)           & $93.0$ / $95.7$ & $0.87$ / $0.92$ \\
PageIndex (Plain / MAVS)         & $97.3$ / $97.7$ & $0.95$ / $0.96$ \\
BGE-M3$+$ann.\ (Plain / MAVS)    & $93.7$ / $96.3$ & $0.89$ / $0.93$ \\
\methodR{} (Plain / MAVS)        & $92.7$ / $94.7$ & $0.87$ / $0.91$ \\
ColPali$+$ann.\ (Plain / MAVS)   & $93.0$ / $95.7$ & $0.88$ / $0.92$ \\
\midrule
\textbf{Mean (14 pipelines)} & \textbf{95.1\%} & \textbf{0.913} \\
Range                        & $92.7$--$97.7$  & $0.87$--$0.96$ \\
\bottomrule
\end{tabular}
\caption{Inter-judge agreement and Cohen's $\kappa$ between GPT-4.1 and
Claude on the same 300-QA outputs. Mean $\kappa{=}0.913$ corresponds to
``almost perfect agreement'' on the Landis-Koch scale. The pipeline-level
ranking under the two judges is identical (ColPali $+$ ann.\ $+$ rk
$+$ \methodS{} is best under both; \methodR{} $+$ \methodS{} second under
Claude, tied second under GPT). This indicates that the observed end-to-end ranking is
not dependent on a single judge's quirks.}
\label{tab:inter-judge}
\end{table}

We do not perform human evaluation due to scale ($14 \times 300 = 4{,}200$
prediction-gold pairs); we view the high inter-judge agreement and the
agreement of pipeline rankings between two independently developed LLMs
as a reasonable proxy.

\section{Extended Limitations Discussion}
\label{app:limits-extended}

We elaborate the limitations summarized in \S\ref{sec:limitations}.

\paragraph{Reranker LLM dependence.}
The rerank stage in this work uses GPT-4.1 at $\tau{=}0$. We expect the
annotation-conditioned rerank protocol to be reranker-agnostic in principle,
because the candidate JSON exposes per-page evidence in a form that any
instruction-tuned LLM can consume. A pilot replacement with a smaller
reranker would falsify or strengthen this expectation; we leave a full
small-reranker / open-source-reranker sweep for follow-up work due to
budget constraints.

\paragraph{LongDocURL coverage.}
LongDocURL serves as an external transfer check: the retrieval-side
comparison (Table~\ref{tab:longdocurl}) reports adaptive-$K$ methods
on the full $N{=}1{,}122$ QAs with non-empty gold evidence, but the
full five-pool cross-pool isolation conducted on MMLongBench-Doc is
not separately replicated on LongDocURL. End-to-end LongDocURL
results (Appendix~\ref{app:full-e2e},
Table~\ref{tab:longdocurl-e2e-full}) confirm the broad pattern.

\paragraph{\methodS{} on extractive workloads.}
On LongDocURL's MCQ-style and short-extractive questions, the topical,
entity, and structural lenses converge on the same extracted answer; the
synthesis call therefore contributes mean $\Delta \approx 0$ across four
LongDocURL pipelines (Appendix~\ref{app:full-e2e},
Table~\ref{tab:longdocurl-e2e-full}). This motivates the on-demand
deployment regime described in \S\ref{sec:method-gen}: \methodS{} is
gated by question format.

\paragraph{Parser substitution.}
We use the PageIndex API as a structural parser for the section-path
field while also benchmarking against the PageIndex retrieval system; the
two roles are independent. Any public PDF parser that yields a section
tree (PyMuPDF, GROBID) could fill this role. A parser-substitution
robustness study is left to future work.

\end{document}